\documentstyle[ijcai95]{article}
\documentstyle{fleqn}

\input{psfig}
\newcommand{\Ccal}{{\cal C}}

\newcommand{\isa}{\mbox{\sc is-a}}
\newcommand{\wclass}[1]{{\sc #1}}
\newcommand{\Prob}{{\rm p}}

\newcommand{\mx}[1]{\raisebox{-1.5ex}{\shortstack{{\rm max}\\{\small $#1$}}}}
\newcommand{\mn}[1]{\raisebox{-1.5ex}{\shortstack{{\rm min}\\{\small $#1$}}}}
\newcommand{\ignore}[1]{}

\title{Using Information Content to Evaluate Semantic Similarity in a Taxonomy}
\author{Philip Resnik\thanks{Appears in Proceedings of
IJCAI-95. Portions of this research 
were supported by an IBM Graduate Fellowship
and grants ARO DAAL 03-89-C-0031, DARPA N00014-90-J-1863, NSF IRI 90-16592,
and Ben Franklin 91S.3078C-1.} \\
Sun Microsystems Laboratories \\
Two Elizabeth Drive \\
Chelmsford, MA 01824-4195 USA \\
{\tt philip.resnik@east.sun.com}
}

\begin{document}

\maketitle 

\begin{abstract}
This paper presents a new measure of semantic similarity in an {\sc is-a}
taxonomy, based on the notion of information content.  Experimental evaluation
suggests that the measure performs encouragingly well (a correlation of $r =
0.79$ with a benchmark set of human similarity judgments, with an upper bound
of $r = 0.90$ for human subjects performing the same task), and significantly
better than the traditional edge counting approach ($r = 0.66$).
\end{abstract}

\section{Introduction}
\label{sec:introduction}

Evaluating semantic relatedness using network representations is a problem
with a long history in artificial intelligence and psychology, dating back to
the spreading activation approach of Quillian \shortcite{quillian1968} and
Collins and Loftus \shortcite{collins1975}.  Semantic similarity represents a
special case of semantic relatedness: for example, cars and gasoline would
seem to be more closely related than, say, cars and bicycles, but the latter
pair are certainly more similar.  Rada et al. \shortcite{Rada89b} suggest that
the assessment of similarity in semantic networks can in fact be thought of as
involving just taxonomic ({\sc is-a}) links, to the exclusion of other link
types; that view will also be taken here, although admittedly it excludes some
potentially useful information.

A natural way to evaluate semantic similarity in a taxonomy is to evaluate the
distance between the nodes corresponding to the items being compared --- the
shorter the path from one node to another, the more similar they are.  Given
multiple paths, one takes the length of the shortest one
\cite{Lee93,Rada89a,Rada89b}.

A widely acknowledged problem with this approach, however, is that it relies
on the notion that links in the taxonomy represent uniform distances.
Unfortunately, this is difficult to define, much less to control.  In real
taxonomies, there is wide variability in the ``distance'' covered by a single
taxonomic link, particularly when certain sub-taxonomies (e.g. biological
categories) are much denser than others.  For example, in WordNet
\cite{miller1990}, a broad-coverage semantic network for English 
constructed by George Miller and colleagues at Princeton, it is not at all
difficult to find links that cover an intuitively narrow distance ({\sc
rabbit ears} {\sc is-a} {\sc television antenna}) or an intuitively wide one
({\sc phytoplankton} {\sc is-a} {\sc living thing}).  The same kinds of
examples can be found in the Collins COBUILD Dictionary \cite{sinc1987a},
which identifies superordinate terms for many words (e.g.  {\sc safety valve}
{\sc is-a} {\sc valve} seems a lot narrower than {\sc knitting machine} {\sc
is-a} {\sc machine}).

In this paper, I describe an alternative way to evaluate semantic similarity
in a taxonomy, based on the notion of information content.  Like the edge
counting method, it is conceptually quite simple.  However, it is not
sensitive to the problem of varying link distances.  In addition, by combining
a taxonomic structure with empirical probability estimates, it provides a way
of adapting a static knowledge structure to multiple contexts.
Section~\ref{sec:proposal} sets up the probabilistic framework and defines the
measure of semantic similarity in information-theoretic terms;
Section~\ref{sec:evaluation} presents an evaluation of the similarity measure
against human similarity judgments, using the simple edge-counting method as a
baseline; and Section~\ref{sec:related} discusses related work.

\section{Similarity and Information Content}
\label{sec:proposal}

Let $\Ccal$ be the set of concepts in an {\sc is-a} taxonomy, permitting
multiple inheritance.
\begin{figure}
\hbox{
\centerline{
\psfig{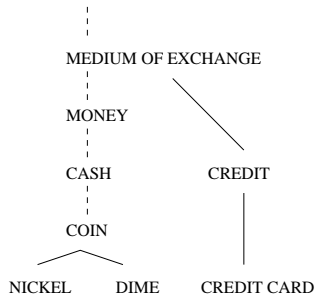}}
}
\caption{Fragment of the WordNet taxonomy.  Solid lines represent {\sc is-a}
links; dashed lines indicate that some intervening nodes were omitted to
save space. \label{fig:taxonomy}}
\end{figure}
Intuitively, one key to the similarity of two concepts is the extent to which
they share information in common, indicated in an {\sc is-a} taxonomy by a
highly specific concept that subsumes them both.  The edge counting method
captures this indirectly, since if the minimal path of {\sc is-a} links
between two nodes is long, that means it is necessary to go high in the
taxonomy, to more abstract concepts, in order to find a least upper bound.
For example, in WordNet, \wclass{nickel} and \wclass{dime} are both subsumed
by \wclass{coin}, whereas the most specific superclass that \wclass{nickel}
and \wclass{credit card} share is \wclass{medium of exchange}.\footnote{In a
feature-based setting (e.g. \cite{tversky1977}), this would be reflected by
explicit shared features: nickels and dimes are both small, round, metallic,
and so on.  These features are captured implicitly by the taxonomy in
categorizing \wclass{nickel} and \wclass{dime} as subordinates of
\wclass{coin}.} (See Figure~\ref{fig:taxonomy}.)

By associating probabilities with concepts in the taxonomy, it is possible to
capture the same idea, but avoiding the unreliability of edge distances.  Let
the taxonomy be augmented with a function $\Prob : \Ccal \rightarrow [0,1]$,
such that for any $c \in \Ccal$, $\Prob(c)$ is the probability of encountering
an instance of concept $c$.  This implies that $\Prob$ is monotonic as one
moves up the taxonomy: if $c_1 \:\isa\: c_2$, then $\Prob(c_1) \leq
\Prob(c_2)$.  Moreover, if the taxonomy has a unique top node then its
probability is 1.  

Following the standard argumentation of information theory
\cite{ross1976}, the {\em information content\/} of a 
concept $c$ can be quantified as negative the log likelihood, $-\log\Prob(c)$.
Notice that quantifying information content in this way makes intuitive sense
in this setting: as probability increases, informativeness decreases, so the
more abstract a concept, the lower its information content.  Moreover, if
there is a unique top concept, its information content is 0.

\begin{figure}
\hbox{
\centerline{
\psfig{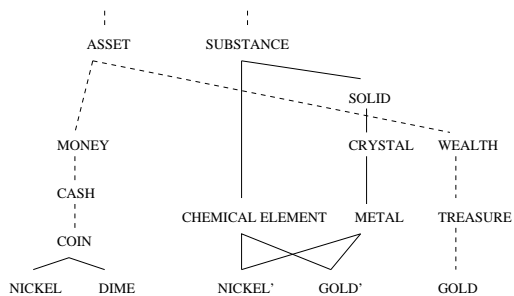}}
}
\caption{Another fragment of the WordNet taxonomy \label{fig:multiple}}
\end{figure}

This quantitative characterization of information provides a new way to
measure semantic similarity.  The more information two concepts share in
common, the more similar they are, and the information shared by two concepts
is indicated by the information content of the concepts that subsume them in
the taxonomy.   Formally, define
\begin{eqnarray}
\mbox{\rm sim}(c_1, c_2) & = &  \mx{c \in S(c_1,c_2)} 
				   \left[ -\log\Prob(c) \right],
\end{eqnarray}
where $S(c_1,c_2)$ is the set of concepts that subsume both $c_1$ and $c_2$.
Notice that although similarity is computed by considering all upper bounds
for the two concepts, the information measure has the effect of identifying
minimal upper bounds, since no class is less informative than its
superordinates.  For example, in Figure~\ref{fig:taxonomy}, \wclass{coin},
\wclass{cash}, etc. are all members of
 $S(\mbox{\wclass{nickel}},\mbox{\wclass{dime}})$, but the concept that is
structurally the minimal upper bound, \wclass{coin}, will also be the most
informative.  This can make a difference in cases of multiple inheritance; for
example, in Figure~\ref{fig:multiple}, \wclass{metal} and
\wclass{chemical element} are not structurally distinguishable as 
upper bounds of \wclass{nickel'} and \wclass{gold'}, but their information
content may in fact be quite different.

In practice, one often needs to measure word similarity, rather than concept
similarity.  Using $s(w)$ to represent the set of concepts in the taxonomy
that are senses of word $w$, define
\begin{eqnarray}
\mbox{\rm sim}(w_1, w_2) & = &  
  \mx{c1,c2} \left[ \mbox{\rm sim}(c_1,c_2) \right],
\end{eqnarray}
where $c_1$ ranges over $s(w_1)$ and $c_2$ ranges over $s(w_2)$.  This is
consistent with Rada et al.'s \shortcite{Rada89b} treatment of ``disjunctive
concepts'' using edge counting: they define the distance between two
disjunctive sets of concepts as the minimum path length from any element of
the first set to any element of the second.  Here, the word similarity is
judged by taking the maximal information content over all concepts of which
both words could be an instance.
For example, Figure~\ref{fig:multiple} illustrates how the similarity of words
{\em nickel\/} and {\em gold\/} would be computed: the information content
would be computed for all classes subsuming any pair in the cross product of
\{\wclass{nickel},\wclass{nickel'}\} and \{\wclass{gold},\wclass{gold'}\}, and
the information content of the most informative class used to quantify the
similarity of the two words.

\section{Evaluation}
\label{sec:evaluation}

\subsection{Implementation}

The work reported here used WordNet's (50,000-node) taxonomy of concepts
represented by nouns (and compound nominals) in English.\footnote{{\em
Concept} as used here refers to what Miller et al. \shortcite{miller1990} call
a {\em synset}, essentially a node in the taxonomy.} Frequencies of concepts
in the taxonomy were estimated using noun frequencies from the Brown Corpus of
American English \cite{francis1982}, a large (1,000,000 word) collection of
text across genres ranging from news articles to science fiction.  Each noun
that occurred in the corpus was counted as an occurrence of each taxonomic
class containing it.\footnote{Plural nouns counted as instances of their
singular forms.} For example, in Figure~\ref{fig:taxonomy}, an occurrence of
the noun {\em dime} would be counted toward the frequency of \wclass{dime},
\wclass{coin}, and so forth.  Formally,
\begin{eqnarray}
\mbox{\rm freq}(c) & = & \sum_{n \in \mbox{words}(c)} \mbox{\rm count}(n),
\end{eqnarray}
where $\mbox{words}(c)$ is the set of words subsumed by concept $c$.  Concept
probabilities were computed simply as relative frequency:
\begin{eqnarray}
\hat{\Prob}(c) & = & \frac{\mbox{\rm freq}(c)}{N},
\end{eqnarray}
where $N$ was the total number of nouns observed (excluding those not subsumed
by any WordNet class, of course).

\subsection{Task}

Although there is no standard way to evaluate computational measures of
semantic similarity, one reasonable way to judge would seem to be agreement
with human similarity ratings.  This can be assessed by using a computational
similarity measure to rate the similarity of a set of word pairs, and looking
at how well its ratings correlate with human ratings of the same pairs.

An experiment by Miller and Charles \shortcite{miller1991:similarity} provided
appropriate human subject data for the task.  In their study, 38 undergraduate
subjects were given 30 pairs of nouns that were chosen to cover high,
intermediate, and low levels of similarity (as determined using a previous
study \cite{rubenstein1965}), and asked to rate ``similarity of meaning'' for
each pair on a scale from 0~(no similarity) to 4~(perfect synonymy).  The
average rating for each pair thus represents a good estimate of how similar
the two words are, according to human judgments.

In order to get a baseline for comparison, I replicated Miller and Charles's
experiment, giving ten subjects the same 30 noun pairs.  The subjects were all
computer science graduate students or postdocs at the University of
Pennsylvania, and the instructions were exactly the same as used by Miller and
Charles, the main difference being that in this replication the subjects
completed the questionnaire by electronic mail (though they were instructed to
complete the whole thing in a single uninterrupted sitting).  Five subjects
received the list of word pairs in a random order, and the other five received
the list in the reverse order.  The correlation between the Miller and Charles
mean ratings and the mean ratings in my replication was .96, quite close to
the .97 correlation that Miller and Charles obtained between their results and
the ratings determined by the earlier study.

For each subject in my replication, I computed how well his or her ratings
correlated with the Miller and Charles ratings.  The average correlation over
the 10 subjects was $r = 0.8848$, with a standard deviation of
0.08.\footnote{Inter-subject correlation in the replication, estimated using
leaving-one-out resampling \cite{weiss1991}, was $r = .9026, {\rm stdev} =
0.07$.} This value represents an upper bound on what one should expect from a
computational attempt to perform the same task.

For purposes of evaluation, three computational similarity measures were used.
The first is the similarity measurement using information content proposed in
the previous section.  The second is a variant on the edge counting method,
converting it from distance to similarity by subtracting the path length from
the maximum possible path length:
\begin{equation}
\mbox{\rm sim}_{\mbox{edge}}(w_1, w_2) = \\
(2 \times \mbox{\sc max}) - \left[ \mn{c_1,c_2}\mbox{len}(c_1,c_2) \right]
 \label{eq:simedge}
\end{equation}
where $c_1$ ranges over $s(w_1)$, $c_2$ ranges over $s(w_2)$, {\sc max} is the
maximum depth of the taxonomy, and $\mbox{len}(c_1,c_2)$ is the length of the
shortest path from $c_1$ to $c_2$ . (Recall that $s(w)$ denotes the set of
concepts in the taxonomy that represent senses of word $w$.)  Note that the
conversion from a distance to a similarity can be viewed as an expository
convenience, and does not affect the evaluation: although the sign of the
correlation coefficient changes from positive to negative, its magnitude turns
out to be just the same regardless of whether or not the minimum path length
is subtracted from $(2 \times \mbox{\sc max})$.

The third point of comparison is a measure that simply uses the probability of
a concept, rather than the information content:
\begin{eqnarray}
\mbox{\rm sim}_{\Prob(c)}(c_1, c_2) & = &  \mx{c \in S(c_1,c_2)} 
				   \left[ 1-\Prob(c) \right] \\
\mbox{\rm sim}_{\Prob(c)}(w_1, w_2) & = &  
			\mx{c_1,c_2}  
			  \left[ \mbox{\rm sim}_{\Prob(c)}(c_1,c_2) \right],
  \label{eq:probsim}
\end{eqnarray}
where $c_1$ ranges over $s(w_1)$ and $c_2$ ranges over $s(w_2)$
in~(\ref{eq:probsim}).  Again, the difference between maximizing $1 -
\Prob(c)$ and minimizing $\Prob(c)$ turns out not to affect the magnitude of
the correlation.  It simply ensures that the value can be interpreted as a
similarity value, with high values indicating similar words.

\subsection{Results}

Table~\ref{tbl:results} summarizes the experimental results, giving the
correlation between the similarity ratings and the mean ratings reported by
Miller and Charles.  Note that, owing to a noun missing from the WordNet
taxonomy, it was only possible to obtain computational similarity ratings for
28 of the 30 noun pairs; hence the proper point of comparison for human
judgments is not the correlation over all 30 items ($r = .8848$), but rather
the correlation over the 28 included pairs ($r = .9015$).
\begin{table}
\begin{center}
\begin{tabular}{|l|r|} \hline
Similarity method 		& Correlation 	\\ \hline\hline
Human judgments	(replication)	& $r = .9015$	\\ \hline
Information content		& $r = .7911$	\\ \hline
Probability			& $r = .6671$	\\ \hline
Edge counting			& $r = .6645$	\\ \hline
\end{tabular}
\end{center}
\caption{Summary of experimental results. \label{tbl:results}}
\end{table}
The similarity ratings by item are given in Table~\ref{tbl:ratings}.

\subsection{Discussion}
\label{sec:discussion}

The experimental results in the previous section suggest that measuring
semantic similarity using information content provides quite reasonable
results, significantly better than the traditional method of simply counting
the number of intervening {\sc is-a} links.  

The measure is not without its problems, however.  One problem is that, like
simple edge counting, the measure sometimes produces spuriously high
similarity measures for words on the basis of inappropriate word senses.  For
example, Table~\ref{tbl:tobacco} shows the word similarity for several words
with {\em tobacco}.  {\em Tobacco\/} and {\em alcohol\/} are similar, both
being drugs, and {\em tobacco\/} and {\em sugar\/} are less similar, though
not entirely dissimilar, since both can be classified as substances.
\begin{table}
\begin{center}
\begin{tabular}{|l|l|r|r|} \hline
n1 & n2 & sim(n1,n2) & class \\ 
\hline\hline
tobacco & alcohol & 7.63  &  \wclass{drug} \\ \hline
tobacco & sugar   & 3.56  &  \wclass{substance} \\ \hline
tobacco & horse   & 8.26  &  \wclass{narcotic} \\ \hline
\end{tabular}
\end{center}
\caption{Similarity with {\em tobacco} computed by maximizing information 
	content
		\label{tbl:tobacco}}
\end{table}
The problem arises, however, in the similarity rating for {\em tobacco\/} with
{\em horse}: the word {\em horse\/} can be used as a slang term for {\em
heroin}, and as a result information-based similarity is maximized, and path
length minimized, when the two words are both categorized as narcotics.  This
is contrary to intuition.

Cases like this are probably relatively rare. However, the example illustrates
a more general concern: in measuring similarity between words, it is really
the relationship among word {\em senses\/} that matters, and a similarity
measure should be able to take this into account.

In the absence of a reliable algorithm for choosing the appropriate word
senses, the most straightforward way to do so in the information-based setting
is to consider {\em all\/} concepts to which both nouns belong rather than
taking just the single maximally informative class.  This suggests redefining
similarity as follows:
\begin{eqnarray}
{\rm sim}(c_1,c_2) & = & \sum_{i}\alpha(c_i)[ -\log\Prob(c_i) ], 
	\label{eq:simalpha}
\end{eqnarray}
where $\{c_i\}$ is the set of concepts dominating both $c_1$ and $c_2$, as
before, and $\sum_{i} \alpha(c_i) = 1$.  This measure of similarity takes more
information into account than the previous one: rather than relying on the
single concept with {\em maximum\/} information content, it allows {\em
each\/} class to contribute information content according to the value of
$\alpha(c_i)$.  Intuitively, these $\alpha$ values measure relevance --- for
example, $\alpha(\mbox{\wclass{narcotic}})$ might be low in general usage but
high in the context of a newspaper article about drug dealers.  In work on
resolving syntactic ambiguity using semantic information
\cite{resnik1993:arpa}, I have found that local syntactic information can be
used successfully to set values for the $\alpha$.

\section{Related Work}
\label{sec:related}

Although the counting of edges in {\sc is-a} taxonomies seems to be something
many people have tried, there seem to be few published descriptions of
attempts to directly evaluate the effectiveness of this method.  A number of
researchers have attempted to make use of conceptual distance in information
retrieval.  For example, Rada et al. \shortcite{Rada89a,Rada89b} and Lee et
al. \shortcite{Lee93} report experiments using conceptual distance,
implemented using the edge counting metric, as the basis for ranking documents
by their similarity to a query.  Sussna \shortcite{sussna1993} uses semantic
relatedness measured with WordNet in word sense disambiguation, defining a
measure of distance that weights different types of links and also explicitly
takes depth in the taxonomy into account.

The most relevant related work appears in an unpublished manuscript by Leacock
and Chodorow \shortcite{leacock1994:ms}.  They have defined a measure
resembling information content, but using the normalized path length between
the two concepts being compared rather than the probability of a subsuming
concept.  Specifically, they define
\begin{eqnarray}
\mbox{\rm sim}_{\mbox{ndist}}(w_1, w_2) & = &
 - \log \left[ \frac{\mn{c_1,c_2}\mbox{len}(c_1,c_2)}{(2\times\mbox{\sc max})}
        \right].
 \label{eq:leacock}
\end{eqnarray}
(The notation above is the same as for equation~(\ref{eq:simedge}).)  In
addition to this definition, they also include several special cases, most
notably to avoid infinite similarity when $c_1$ and $c_2$ are exact synonyms
and thus have a path length of 0. Leacock and Chodorow have experimented with
this measure and the information content measure described here in the context
of word sense disambiguation, and found that they yield roughly similar
results.  More significantly, I recently implemented their method and tested
it on the task reported in the previous section, and found that it actually
outperforms the information-based measure.  This led me to do a followup
experiment using a different and larger set of noun pairs, and in the followup
study the information-based measure performed better.\footnote{In the followup
study, I used netnews archives to gather highly frequent nouns within related
topic areas, and then selected noun pairings at random, in order to avoid
biasing the followup study in favor of either algorithm.} The relationship
between the two algorithms will thus require further study.  For now, however,
what seems most significant is that both approaches take the form of a
log-based (and hence information-like) measure, as originally proposed in
\cite{resnik:dissertation}.

Finally, in the context of current research in computational linguistics, the
approach to semantic similarity taken here can be viewed as a hybrid,
combining corpus-based statistical methods with knowledge-based taxonomic
information.  The use of corpus statistics alone in evaluating word similarity
--- without prior taxonomic knowledge --- is currently an active area of
research in the natural language community.  This is largely a reaction to
sparse data problems in training statistical language models: it is difficult
to come up with an accurate statistical characterization of the behavior of
words that have been encountered few times or not at all.  Word similarity
appears to be one promising way to solve the problem: the behavior of a word
is approximated by smoothing its observed behavior together with the behavior
of words to which it is similar.  For example, a speech recognizer that has
never seen the phrase {\em ate a peach\/} can still conclude that {\em John
ate a peach\/} is a reasonable sequence of words in English if it has seen
other sentences like {\em Mary ate a pear\/} and knows that {\em peach\/} and
{\em pear\/} have similar behavior.

The literature on corpus-based determination of word similarity has recently
been growing by leaps and bounds, and is too extensive to discuss in detail
here (for a review, see \cite{resnik:dissertation}), but most approaches to
the problem share a common assumption: semantically similar words have similar
distributional behavior in a corpus.  Using this assumption, it is common to
treat the words that co-occur near a word as constituting features, and to
compute word similarity in terms of how similar their feature sets are.  As in
information retrieval, the ``feature'' representation of a word often takes
the form of a vector, with the similarity computation amounting to a
computation of distance in a highly multidimensional space.  Given a distance
measure, it is not uncommon to derive word classes by hierarchical clustering.
A difficulty with most distributional methods, however, is how the measure of
similarity (or distance) is to be interpreted.  Although word classes
resulting from distributional clustering are often described as ``semantic,''
they often capture syntactic, pragmatic, or stylistic factors as well.

\section{Conclusions}
\label{sec:conclusions}

This paper has presented a new measure of semantic similarity in an {\sc is-a}
taxonomy, based on the notion of information content.  Experimental evaluation
was performed using a large, independently constructed corpus, an
independently constructed taxonomy, and previously existing human subject
data.  The results suggest that the measure performs encouragingly well (a
correlation of $r = 0.79$ with a benchmark set of human similarity judgments,
against an upper bound of $r = 0.90$ for human subjects performing the same
task), and significantly better than the traditional edge counting approach
($r = 0.66$).

In ongoing work, I am currently exploring the application of
taxonomically-based semantic similarity in the disambiguation of word senses
\cite{resnik1995:wvlc3}.  The idea behind the approach is that when polysemous
words appear together, the appropriate word senses to assign are often those
that share elements of meaning.  Thus {\em doctor\/} can refer to either a
Ph.D. or an M.D., and {\em nurse\/} can signify either a health professional
or someone who takes care of small children; but when {\em doctor\/} and {\em
nurse\/} are seen together, the Ph.D. sense and the childcare sense go by the
wayside.  In a widely known paper, Lesk \shortcite{lesk1986} exploits
dictionary definitions to identify shared elements of meaning --- for example,
in the Collins COBUILD Dictionary \cite{sinc1987a}, the word {\em ill\/} can
be found in the definitions of the correct senses.  More recently, Sussna
\shortcite{sussna1993} has explored using similarity of word senses based on
WordNet for the same purpose.  The work I am pursuing is similar in spirit to
Sussna's approach, although the disambiguation algorithm and the similarity
measure differ substantially.

\begin{table*}
\begin{center}
\begin{tabular}{|c|c|r|r|r|r|r|} \hline
\multicolumn{2}{|c|}{Word Pair}	& Miller and Charles & Replication 
  & sim
  & $\mbox{\rm sim}_{\mbox{edge}}$
  & $\mbox{\rm sim}_{\Prob(c)}$ \\
  &  & means & means & & & 
\\ \hline \hline
car             & automobile      & 3.92    & 3.9   &   8.0411 & 30 & 0.9962 \\
gem             & jewel           & 3.84    & 3.5   &  14.9286 & 30 & 1.0000 \\
journey         & voyage          & 3.84    & 3.5   &   6.7537 & 29 & 0.9907 \\
boy             & lad             & 3.76    & 3.5   &   8.4240 & 29 & 0.9971 \\
coast           & shore           & 3.70    & 3.5   &  10.8076 & 29 & 0.9994 \\
asylum          & madhouse        & 3.61    & 3.6   &  15.6656 & 29 & 1.0000 \\
magician        & wizard          & 3.50    & 3.5   &  13.6656 & 30 & 0.9999 \\
midday          & noon            & 3.42    & 3.6   &  12.3925 & 30 & 0.9998 \\
furnace         & stove           & 3.11    & 2.6   &   1.7135 & 23 & 0.6951 \\
food            & fruit           & 3.08    & 2.1   &   5.0076 & 27 & 0.9689 \\
bird            & cock            & 3.05    & 2.2   &   9.3139 & 29 & 0.9984 \\
bird            & crane           & 2.97    & 2.1   &   9.3139 & 27 & 0.9984 \\
tool            & implement       & 2.95    & 3.4   &   6.0787 & 29 & 0.9852 \\
brother         & monk            & 2.82    & 2.4   &   2.9683 & 24 & 0.8722 \\
crane           & implement       & 1.68    & 0.3   &   2.9683 & 24 & 0.8722 \\
lad             & brother         & 1.66    & 1.2   &   2.9355 & 26 & 0.8693 \\
journey         & car             & 1.16    & 0.7   &   0.0000 & 0  & 0.0000 \\
monk            & oracle          & 1.10    & 0.8   &   2.9683 & 24 & 0.8722 \\
food            & rooster         & 0.89    & 1.1   &   1.0105 & 18 & 0.5036 \\
coast           & hill            & 0.87    & 0.7   &   6.2344 & 26 & 0.9867 \\
forest          & graveyard       & 0.84    & 0.6   &   0.0000 & 0  & 0.0000 \\
monk            & slave           & 0.55    & 0.7   &   2.9683 & 27 & 0.8722 \\
coast           & forest          & 0.42    & 0.6   &   0.0000 & 0  & 0.0000 \\
lad             & wizard          & 0.42    & 0.7   &   2.9683 & 26 & 0.8722 \\
chord           & smile           & 0.13    & 0.1   &   2.3544 & 20 & 0.8044 \\
glass           & magician        & 0.11    & 0.1   &   1.0105 & 22 & 0.5036 \\
noon            & string          & 0.08    & 0.0   &   0.0000 & 0  & 0.0000 \\
rooster         & voyage          & 0.08    & 0.0   &   0.0000 & 0  & 0.0000 \\
\hline
\end{tabular}
\end{center}
\caption{Semantic similarity by item. \label{tbl:ratings}}
\end{table*}

%%% Bibliography goes here, still in two-column mode
% \bibliography{general,learning,distrib,nlstat,ibm_master,senses} 

\begin{thebibliography}{}

\bibitem[\protect\citeauthoryear{Collins and Loftus}{1975}]{collins1975}
A.~Collins and E.~Loftus.
\newblock A spreading activation theory of semantic processing.
\newblock {\em Psychological Review}, 82:407--428, 1975.

\bibitem[\protect\citeauthoryear{Francis and Ku\v{c}era}{1982}]{francis1982}
W.~N. Francis and H.~Ku\v{c}era.
\newblock {\em Frequency Analysis of English Usage: Lexicon and Grammar}.
\newblock Houghton Mifflin, 1982.

\bibitem[\protect\citeauthoryear{Leacock and Chodorow}{1994}]{leacock1994:ms}
Claudia Leacock and Martin Chodorow.
\newblock Filling in a sparse training space for word sense identification.
\newblock ms., March 1994.

\bibitem[\protect\citeauthoryear{Lee \bgroup \em et al.\egroup }{1993}]{Lee93}
Joon~Ho Lee, Myoung~Ho Kim, and Yoon~Joon Lee.
\newblock Information retrieval based on conceptual distance in {IS-A}
  hierarchies.
\newblock {\em Journal of Documentation}, 49(2):188--207, June 1993.

\bibitem[\protect\citeauthoryear{Lesk}{1986}]{lesk1986}
Michael Lesk.
\newblock Automatic sense disambiguation using machine readable dictionaries:
  how to tell a pine cone from an ice cream cone.
\newblock In {\em Proceedings of the 1986 SIGDOC Conference}, pages 24--26,
  1986.

\bibitem[\protect\citeauthoryear{Miller and
  Charles}{1991}]{miller1991:similarity}
George~A. Miller and Walter~G. Charles.
\newblock Contextual correlates of semantic similarity.
\newblock {\em Language and Cognitive Processes}, 6(1):1--28, 1991.

\bibitem[\protect\citeauthoryear{Miller}{1990}]{miller1990}
George Miller.
\newblock {WordNet}: An on-line lexical database.
\newblock {\em International Journal of Lexicography}, 3(4), 1990.
\newblock (Special Issue).

\bibitem[\protect\citeauthoryear{Quillian}{1968}]{quillian1968}
M.~Ross Quillian.
\newblock Semantic memory.
\newblock In M.~Minsky, editor, {\em Semantic Information Processing}. MIT
  Press, Cambridge, MA, 1968.

\bibitem[\protect\citeauthoryear{Rada and Bicknell}{1989}]{Rada89a}
Roy Rada and Ellen Bicknell.
\newblock Ranking documents with a thesaurus.
\newblock {\em JASIS}, 40(5):304--310, September 1989.

\bibitem[\protect\citeauthoryear{Rada \bgroup \em et al.\egroup
  }{1989}]{Rada89b}
Roy Rada, Hafedh Mili, Ellen Bicknell, and Maria Blettner.
\newblock Development and application of a metric on semantic nets.
\newblock {\em {IEEE} Transaction on Systems, Man, and Cybernetics},
  19(1):17--30, February 1989.

\bibitem[\protect\citeauthoryear{Resnik}{1993a}]{resnik:dissertation}
Philip Resnik.
\newblock {\em Selection and Information: A Class-Based Approach to Lexical
  Relationships}.
\newblock PhD thesis, University of Pennsylvania, December 1993.

\bibitem[\protect\citeauthoryear{Resnik}{1993b}]{resnik1993:arpa}
Philip Resnik.
\newblock Semantic classes and syntactic ambiguity.
\newblock In {\em Proceedings of the 1993 {ARPA} Human Language Technology
  Workshop}. Morgan Kaufmann, March 1993.

\bibitem[\protect\citeauthoryear{Resnik}{1995}]{resnik1995:wvlc3}
Philip Resnik.
\newblock Disambiguating noun groupings with respect to {WordNet} senses.
\newblock In {\em Third Workshop on Very Large Corpora}. Association for
  Computational Linguistics, 1995.

\bibitem[\protect\citeauthoryear{Ross}{1976}]{ross1976}
Sheldon Ross.
\newblock {\em A First Course in Probability}.
\newblock Macmillan, 1976.

\bibitem[\protect\citeauthoryear{Rubenstein and
  Goodenough}{1965}]{rubenstein1965}
Herbert Rubenstein and John Goodenough.
\newblock Contextual correlates of synonymy.
\newblock {\em CACM}, 8(10):627--633, October 1965.

\bibitem[\protect\citeauthoryear{Sinclair~(ed.)}{1987}]{sinc1987a}
John Sinclair~(ed.).
\newblock {\em Collins {COBUILD} English Language Dictionary}.
\newblock Collins: London, 1987.

\bibitem[\protect\citeauthoryear{Sussna}{1993}]{sussna1993}
Michael Sussna.
\newblock Word sense disambiguation for free-text indexing using a massive
  semantic network.
\newblock In {\em Proceedings of the Second International Conference on
  Information and Knowledge Management (CIKM-93)}, Arlington, Virginia, 1993.

\bibitem[\protect\citeauthoryear{Tversky}{1977}]{tversky1977}
A.~Tversky.
\newblock Features of similarity.
\newblock {\em Psychological Review}, 84:327--352, 1977.

\bibitem[\protect\citeauthoryear{Weiss and Kulikowski}{1991}]{weiss1991}
Sholom~M. Weiss and Casimir~A. Kulikowski.
\newblock {\em Computer systems that learn: classification and prediction
  methods from statistics, neural nets, machine learning, and expert systems}.
\newblock Morgan Kaufmann, San Mateo, CA, 1991.

\end{thebibliography}
\bibliographystyle{named}

\end{document}